%% file: main.tex
\documentclass{article}
\pdfoutput=1
\usepackage{microtype}
\usepackage{graphicx}
\usepackage{subfigure}
\usepackage{booktabs} 
\usepackage[colorinlistoftodos]{todonotes}

\usepackage{hyperref}

\usepackage{tabularx}
\usepackage{caption}
\usepackage{float}
\usepackage{amsmath}  
\usepackage{amssymb}  
\usepackage{algorithm}  
\usepackage{algpseudocodex}  
\usepackage{accents} 
\usepackage{setspace}  
\usepackage{xcolor} 
\usepackage{bm} 
\usepackage{csquotes}
\usepackage[normalem]{ulem}
\usepackage{capt-of}

\newcommand\blfootnote[1]{
    \begingroup
    \renewcommand\thefootnote{}\footnote{#1}
    \addtocounter{footnote}{-1}
    \endgroup
}


\makeatletter
\patchcmd{\@algocf@start}
  {-5.5em}
  {0pt}
  {}{}
\makeatother

\input{math}

\usepackage[accepted]{mlsys2024}

\mlsystitlerunning{Efficient LLM Inference using Dynamic Input Pruning and Cache-Aware Masking}

\begin{document}

\twocolumn[

\mlsystitle{Efficient LLM Inference using \\Dynamic Input Pruning and Cache-Aware Masking}

\mlsyssetsymbol{equal}{*}

\begin{mlsysauthorlist}
\mlsysauthor{Marco Federici*}{qc}
\mlsysauthor{Davide Belli*}{qc}
\mlsysauthor{Mart van Baalen}{qc}
\mlsysauthor{Amir Jalalirad}{qc}
\mlsysauthor{Andrii Skliar}{qc}
\mlsysauthor{Bence Major}{qc}
\mlsysauthor{Markus Nagel}{qc}
\mlsysauthor{Paul Whatmough}{qc}
\end{mlsysauthorlist}

\mlsysaffiliation{qc}{Qualcomm AI Research. Qualcomm AI Research is an initiative of Qualcomm Technologies, Inc}

\mlsyskeywords{Machine Learning, MLSys, Dynamic Sparsity, Model Efficiency}

\mlsyscorrespondingauthor{Marco Federici}{mfederic@qti.qualcomm.com}
\mlsyscorrespondingauthor{Davide Belli}{dbelli@qti.qualcomm.com}
\mlsyscorrespondingauthor{Mart van Baalen}{mart@qti.qualcomm.com}
\mlsyscorrespondingauthor{Markus Nagel}{markusn@qti.qualcomm.com}

\vskip 0.3in

\begin{abstract}
While mobile devices provide ever more compute power, improvements in DRAM bandwidth are much slower.
This is unfortunate for large language model (LLM) token generation, which is heavily memory-bound.
Previous work has proposed to leverage natural dynamic activation sparsity in ReLU-activated LLMs to reduce effective DRAM bandwidth per token.
However, more recent LLMs use SwiGLU instead of ReLU, which results in little inherent sparsity. While SwiGLU activations can be pruned based on magnitude, the resulting sparsity patterns are difficult to predict, rendering previous approaches ineffective.
To circumvent this issue, our work introduces Dynamic Input Pruning (DIP): a predictor-free dynamic sparsification approach,
which preserves accuracy with minimal fine-tuning.
DIP can further use lightweight LoRA adapters to regain some performance lost during sparsification. 
Lastly, we describe a novel cache-aware masking strategy, which considers the cache state and activation magnitude to further increase cache hit rate, improving LLM token rate on mobile devices.
DIP outperforms other methods in terms of accuracy, memory and throughput trade-offs across simulated hardware settings. 
On Phi-3-Medium, DIP achieves a 46\% reduction in memory and 40\% increase in throughput with $<$ 0.1 loss in perplexity when compared to streaming the dense model from Flash. 
The open source code for HW simulator, methods, and experiments in this paper is available at \href{https://github.com/Qualcomm-AI-research/dynamic-sparsity}{https://github.com/Qualcomm-AI-research/dynamic-sparsity}.
\end{abstract}

]  

\printAffiliationsAndNotice{\mlsysEqualContribution} 

\section{Introduction}
\label{sec:intro}

\blfootnote{Snapdragon branded products are products of Qualcomm Technologies, Inc. and/or its subsidiaries.}

\vspace{-3mm}

\begin{figure}[h!]
    \centering
    \includegraphics[width=0.9\columnwidth]{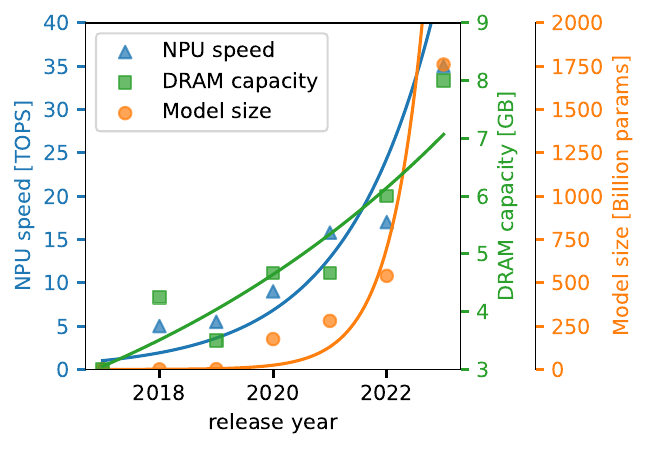}
        \caption{Comparing the largest LLMs by year of release \cite{zhao2023survey} against NPU and DRAM trends in recent iPhone devices\footnotemark. An exponential fit highlights the linear increase in DRAM size opposed to the exponential increases in NPU processing power and model sizes.} 
    \label{fig:trends_dram_npu} 
\end{figure}

Specialized neural processing unit (NPU) accelerators for AI have become standard in recent smartphone SoCs. 
These processors enable efficient neural network inference on-device, opening the way for powerful on-device AI applications. 
However, as shown in Figure~\ref{fig:trends_dram_npu}, memory 
size has not improved at the same pace as processing power \cite{jhang2021challenges, kwon2023efficient, gholami2024ai}. 
This memory bottleneck is especially evident for LLMs, which grew to billions and trillions of parameters, and for which all weights must be accessed to generate each individual output token. 

\begin{figure}[bth!]
    \centering
    \includegraphics[width=0.9\columnwidth]{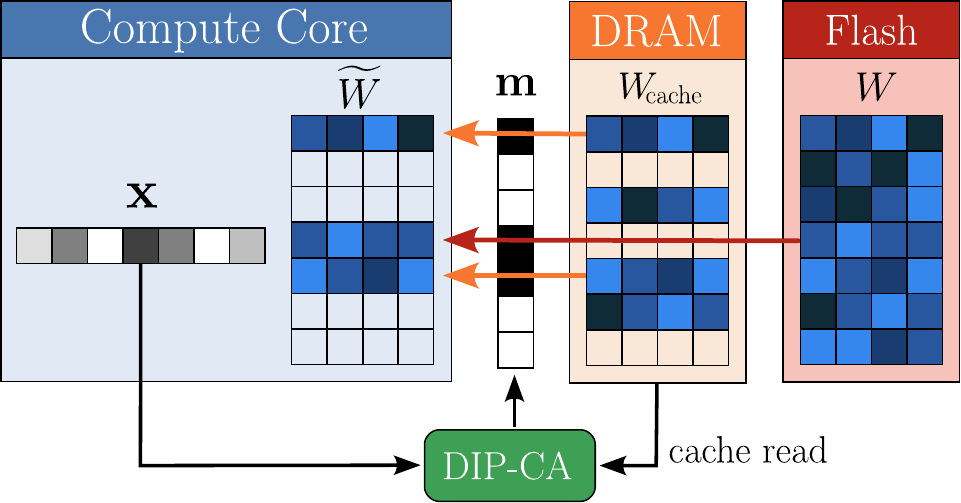}
        \caption{Overview of Cache-Aware Dynamic Input Pruning for a vector-matrix multiplication. The Cache-Aware masking component (DIP-CA) outputs a column sparsity mask based on current input activations and weights currently cached in DRAM. The subset of required weights are retrieved from DRAM cache, if available, or loaded from Flash in case of cache misses.}
    \label{fig:diagram_mlp} 
\end{figure}

For instance, a model with 14B parameters like Phi-3-Medium quantized at INT4 precision occupies approximately 7GB, which approaches the average smartphone \textit{total} DRAM capacity \cite{smartphone2020counterpoint, smartphone2024grossi}.

But, in fact, the OS and active applications already occupy a significant portion of the total DRAM capacity,
typically leaving a few GBs free for an on-device LLM.

To relax the DRAM capacity pressure, it has been previously proposed to load a model directly from flash, instead of first copying it in DRAM~\cite{alizadeh2023llm,xue2024powerinfer}. 
By leveraging \emph{dynamic sparsity} and introducing \emph{DRAM caching} the latency overhead of loading a model from flash can be mitigated significantly.
Dynamic sparsity \cite{mirzadeh2023relu, lee2024cats, akhauri2024shadowllm} exploits the fact that all operations during token generation are matrix-vector multiplications, and the fact that in LLMs with ReLU activations, vectors contain many zero elements. 
For elements in the activation vectors that are zero, we can skip loading the corresponding columns in the following layer.
If sparsity predictors such as those introduced in DejaVu \citep{liu2023deja} are used to predict which elements in the activation vectors will be zero, the weights that \emph{generate} zero-valued activation elements can also be ignored in the preceding layer.
Additionally, a DRAM cache is used to cache active parameters from previous tokens. 
When a DRAM cache is used, parameters only need to be loaded from flash in case of a \emph{cache miss}, i.e., when an active parameter is not already in cache from previous tokens.
These approaches allow even Mixtral8x7B to run on mobile devices at reasonable token generation rates \cite{xue2024powerinfer}.

\footnotetext{Data collected from Wikipedia \yrcite{enwiki_apple}. Values are averaged in case of multiple model releases per year.}  

Prior approaches rely on predictors that can anticipate the natural dynamic activation sparsity.
As we will show in more detail in Section \ref{sec:dynamic_sparsity}, SwiGLU MLP used in most modern LLMs \cite{jiang2023mistral, bai2023qwen, abdin2024phi, dubey2024llama} have little to no natural activation sparsity.
While these activations can still be pruned based on magnitude, it turns out to be much harder to predict which activations should be pruned in SwiGLU LLMs than in ReLU LLMs (see Figure~\ref{fig:difficult_pruning}).
This issue is circumvented in previous works \cite{mirzadeh2023relu, song2024prosparse, song2024turbo} by replacing SiLU activations with ReLU activations, and fine-tuning the LLM to adapt to the architecture change.
However, this fine-tuning step is expensive, often requiring billions of tokens of fine-tuning data to achieve reasonable accuracy.
In this work we introduce Dynamic Input Pruning as a predictor-free alternative to dynamic sparsity that does not rely on expensive re-training. We also show that small LoRA adapters \cite{hu2021lora} can be used to mitigate the performance loss due to dynamic pruning.

The high natural sparsity rates of ReLU LLMs (up to 90\%) have two implications: 1) for each token, only a few active parameters need to be loaded from Flash, and 2) active parameters from multiple previous tokens can be cached, leading to higher cache hit rate and, as a result, reduced flash access.
SwiGLU LLMs, however, can be pruned dynamically only up to approximately 50\% before model accuracy is significantly impacted (see Section \ref{sec:dynamic_sparsity}).
Hence, to reach reasonable token generation rates at lower sparsity levels, we introduce a \emph{cache-aware} variant of dynamic input pruning, which we visualize in Figure \ref{fig:diagram_mlp}.
This variant takes into account the current state of the cache when deciding which parameters to use.
By biasing the parameter selection towards the ones that are already in cache, we can increase cache-hit rate and hence LLM throughput, while minimally affecting model accuracy.

The main contributions in this paper are three-fold:
\begin{enumerate}
    \item We analyze the limitation of methods like DejaVu \cite{liu2023deja} and \enquote{LLM in a Flash} \cite{alizadeh2023llm} in modeling activation sparsity for modern non-ReLU LLMs, demonstrating the impracticality of training predictors for SwiGLU-based models (Section \ref{sec:dynamic_sparsity}). 
    
    \item We propose a predictor-free method to dynamically prune MLP layers which outperforms previous static and dynamic pruning methods on the latest LLMs across various sparsity levels. Additionally, we show that a lightweight LoRA component can further reduce the accuracy lost with sparsification. (Section~\ref{sec:dynamic_input_pruning}).
    
    \item We implement a cache-aware masking strategy which modifies the sparsity mask by prioritizing cached weights, achieving better trade-offs between model accuracy and latency at different DRAM constraints and hardware settings (Section~\ref{sec:method_cache}).
\end{enumerate}


\section{Background and Related Work}
\label{sec:related_work}

\begin{figure}[tb]
    \centering
    \begin{minipage}{0.45\columnwidth}
        \centering
        ~~~~~~~Mistral-7B\\
        \includegraphics[width=\textwidth, trim={0 0 0 0.8cm}, clip]{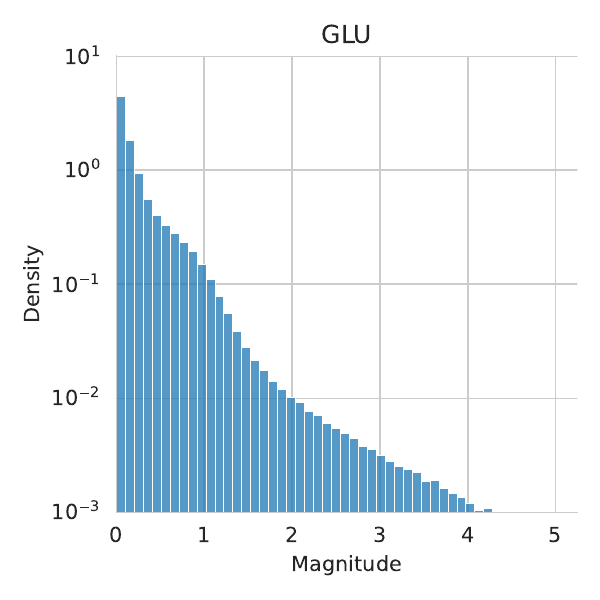}
    \end{minipage}
     \begin{minipage}{0.45\columnwidth}
        \centering
        ~~~~~~Turbosparse-Mistral\\
        \includegraphics[width=\textwidth, trim={0 0 0 0.8cm}, clip]{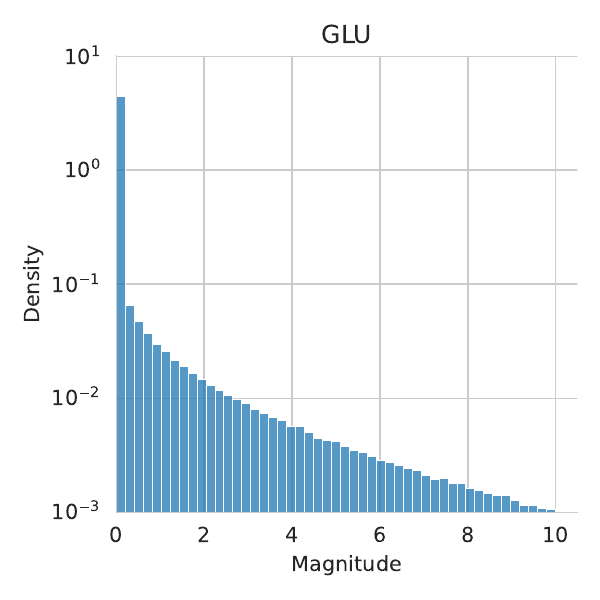}
    \end{minipage}
    \vspace{-0.2cm}
    \caption{GLU Activation Magnitude distribution for Layer 31 Mistral-7B (left), and the corresponding re-trained ReLU-fied Turbosparse Mistral (right).} 
    \label{fig:activation-magnitude}
\end{figure}

\begin{figure}[tb]
    \centering
    \begin{minipage}{0.38\columnwidth}
        \centering
        \footnotesize
        Global\\Threshold\\
    \includegraphics[width=\textwidth,]{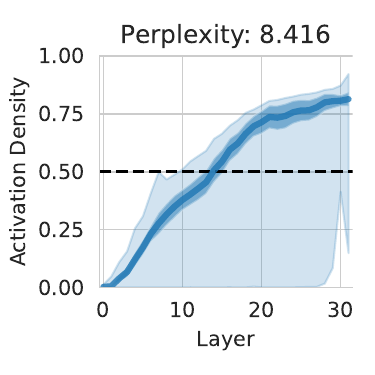}
    \end{minipage}
    \begin{minipage}{0.295\columnwidth}
        \centering
        \footnotesize
        Per-Layer Threshold\\
    \includegraphics[width=\textwidth, clip, trim={1.5cm 0 0 0}]{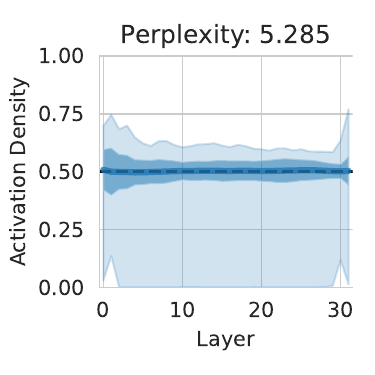}
    \end{minipage}
    \begin{minipage}{0.30\columnwidth}
        \centering
        \footnotesize
        Per-Token Threshold ($\topk$)\\

    \includegraphics[width=\textwidth, trim={1.5cm 0 0 0.8mm}, clip]{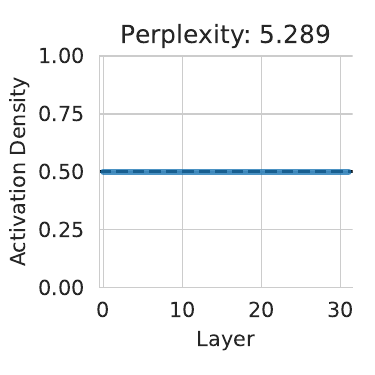}
    \end{minipage}
    \vspace{-0.5cm}
    \caption{Layer activation density for different GLU thresholding strategies at 50\% MLP target density on a Mistral-7B model. Darker shaded region indicate the standard deviation, while lighter shading covers minimum and maximum values. Setting a threshold for each layer or each token result in similar perplexity. The dense model perplexity is $5.249$.}
    \label{fig:threshold-type}
\end{figure}

This section first reviews quantization and static sparsity, two alternative approaches to reduce model size, and later discusses prior works employing dynamic sparsity.

\paragraph{Static Pruning and Quantization}
Pruning and quantization techniques have been extensively explored to minimize the memory footprint of neural networks~\citep{kuzmin2023pruning}. Unstructured pruning typically involves removing individual weight parameters, whereas structured pruning eliminates entire weight structures \citep{van2023llm, frantar2023sparsegpt, sun2023simple}.
Alternative approximation strategies involve low bit-width quantization \citep{nagel2021whitepaper, frantar2022gptq, van2024gptvq, huang2024billm} and low-rank representations \citep{zhang2024oats}. All prior techniques, however, tend to degrade performance, which can be partially mitigated through model fine-tuning. This fine-tuning process necessitates a representative dataset and significant computational resources, and it is therefore prohibitive for certain applications. 
Another downside of static pruning is its limited flexibility, as the set of retained weights is fixed for all activations, while the optimal set of weights may vary across different tokens. 

\paragraph{Dynamic Sparsity}
Recent work proposed dynamic sparsity methods \cite{mirzadeh2023relu, liu2023deja, lee2024cats, akhauri2024shadowllm} to select active weights based on the current inputs, and thus overcoming part of the limitations in static methods.

Recent approaches \cite{mirzadeh2023relu, alizadeh2023llm, song2024turbo} aim to reduce the number of MLP parameters loaded at inference time by exploiting the sparse activation pattern of the intermediate ReLU activations. 
\citet{liu2023deja} (\textit{DejaVu}) further showed that, in LLM with ReLU activations such as OPT \cite{zhang2022opt} and Llama v1 \cite{roumeliotis10llama}, even small predictors can accurately determine active units with moderate training and memory overhead.
Whenever the predictions are correct, the DejaVu approach does not introduce any approximation error since only rows and column corresponding to zero entries are removed.

Other recent work extended the approach from DejaVu aiming to reduce system latency and memory overhead. 
\citet{akhauri2024shadowllm} replaced the individual predictors with a single, global predictor to determine the active units. \citet{alizadeh2023llm} added an LFU cache on the MLP parameters and introduced weight bundling to reduce memory overhead. \citet{song2023powerinfer} proposed identifying most active neurons offline and pre-loading them on GPU for faster access, while loading the remaining neurons on-demand based on predictor outputs. \citet{xue2024powerinfer} builds an engine to optimize LLM inference given certain input hardware specifications. Interestingly, this work shows that neuron pre-loading is not necessary to achieve high system throughput.

Most dynamic sparsity approaches inherently rely on the sparse activation pattern induced by the ReLU non-linearity, however, the vast majority of recent LLM architectures\footnote{ All the open-source LLMs benchmarked in \citet{wang2024mmlu} use SwiGLU non-linearities.}\cite{jiang2023mistral, abdin2024phi, dubey2024llama} use non-linearities that do not naturally induce many hard zeros (cf. Figure~\ref{fig:activation-magnitude}).
While \emph{ReLU-fication}, the process of replacing the original network activations with ReLU activations, has shown promising results \citep{mirzadeh2023relu, song2024turbo}, though on the other hand it requires a computationally demanding retraining procedure which may fail to replicate the original performance \citep{llm2023blog,lee2024cats}.

For this reason, in the following sections we investigate the effectiveness of dynamic sparsity strategies when applied directly to modern LLMs.

\section{Dynamic Sparsity on Modern LLMs}
\label{sec:dynamic_sparsity}

\begin{figure*}[tbh]
    \centering
    \begin{minipage}{0.22\textwidth}
        \centering
        {\small (a) GLU pruning}\\~\\
        \includegraphics[width=\columnwidth]{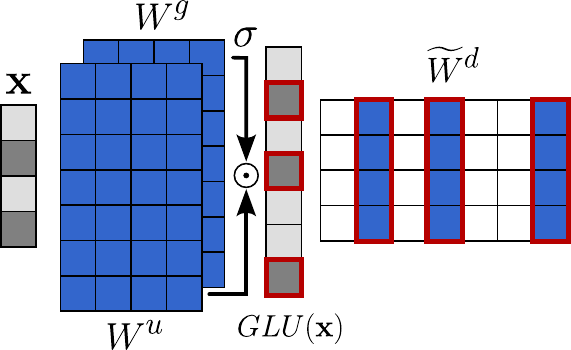}
    \end{minipage}
    \begin{minipage}{0.03\textwidth}
    ~
    \end{minipage}
     \begin{minipage}{0.22\textwidth}
        \centering
        {\small (b) Gate pruning}\\~\\
        \includegraphics[width=\columnwidth]{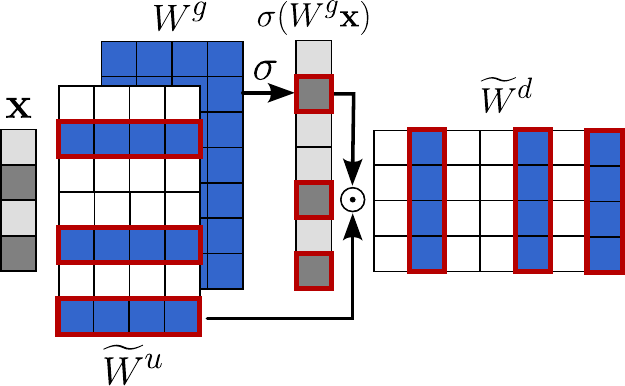}
    \end{minipage}
    \begin{minipage}{0.03\textwidth}
    ~
    \end{minipage}
     \begin{minipage}{0.22\textwidth}
        \centering
        {\small (c) Predictive GLU pruning}\\~\\
        \vspace{-3mm}
        \includegraphics[width=\columnwidth]{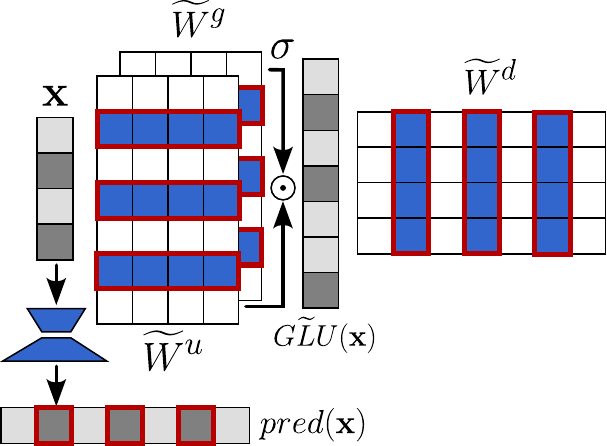}
    \end{minipage}
    \begin{minipage}{0.03\textwidth}
    ~
    \end{minipage}
     \begin{minipage}{0.22\textwidth}
         \centering
        {\small (d) Input pruning (\textbf{ours})}\\~\\
        \includegraphics[width=\columnwidth]{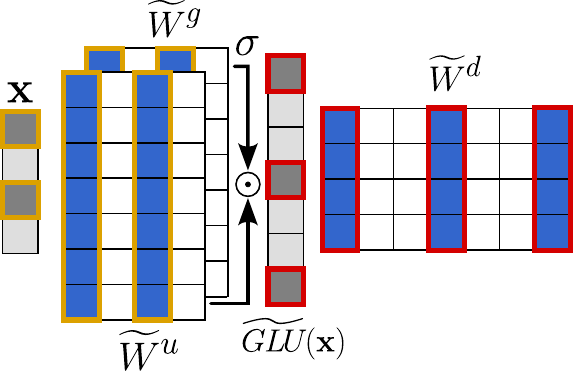}
    \end{minipage}
    \caption{Diagrams describing the sparsity schemes for GLU pruning, Gate pruning, Predictive pruning and Input pruning. Activations with larger magnitude are visualized in darker colors, while dark blue indicates weights used for computation for the current input.} 
    \label{fig:sparsification_diagrams}
\end{figure*}

Modern LLMs consist of alternating sequences of \textit{attention} and \textit{MLP} blocks.
In most cases, the majority of network weights resides in the latter component. This became even more pronounced with the introduction of \textit{Group Query Attention} \citep{roumeliotis10llama}, reducing the number of parameters in attention layers.
MLP architectures in recent LLMs \cite{jiang2023mistral, abdin2024phi, dubey2024llama} frequently consist of a \textit{Gated non-Linear Unit} (GLU) layer, which maps the activations into high-dimensional feature space followed by a linear down projection:
\begin{align}
    \MLP (\rvx) &= \Wd \GLU(\rvx)\\
    \GLU(\rvx) &= \Wu \rvx\odot\sigma(\Wg\rvx)
\end{align}
in which $\Wu$, $\Wg$, $\Wd$ will be referenced as \textit{up}, \textit{gate}, and \textit{down} weight matrices, respectively.
The non-linearity $\sigma$ often consists of a SiLU function \cite{dauphin2017language,elfwing2018sigmoid,naveed2023comprehensive}.

We define as follows the sparse version of an MLP: 
\begin{align}
    \aMLP(\rvx) = \aWd \left(\aWu \rvx\odot \sigma(\aWg\rvx)\right),
\end{align}
where the tilde indicates that the original weight matrices are replaced with a sparse version, which may result in an approximation error.

As discussed in Section~\ref{sec:related_work}, SwiGLU MLPs exhibit no natural activation sparsity, and existing methods to circumvent this are expensive and time-consuming. In the remainder of this section we will first show that non-sparse activations can still be pruned based on magnitude. We then discuss how this can be used to dynamically prune weights. Lastly, we demonstrate that predictor-based approaches such as those introduced by DejaVu \citep{liu2023deja} and its derivatives fail for dynamic, magnitude-based sparsity.

\subsection{Magnitude-based Dynamic Pruning}\label{sec:magnitude_based_pruning}
Since the number of hard zeros in SwiGLU architectures is negligible, to enable higher sparsity  we first relax the condition of removing only zeros, and instead prune values that are smaller than a predetermined threshold $t$.

In Figure~\ref{fig:threshold-type}, we compare the effect of three GLU thresholding strategies on Mistral-7B layer activations in terms of WikiText-2 perplexity: a \emph{global threshold}, i.e., a fixed threshold $t$ is used to prune the activations in all layers; a \emph{per layer} threshold, i.e. a fixed threshold which is set independently per layer based on the CDF of activations over a small calibration set, but shared over different tokens; and a \emph{per-token} threshold, which is based on the CDF of the current activations of each layer dynamically. Note that the latter corresponds to simply keeping the top-K largest magnitude values for each token. In all cases the thresholds are chosen to ensure 50\% of activations are kept on average.

We find that a global threshold performs poorly compared to a per-layer or per-token threshold. This is due to the fact that activation magnitude varies considerably per layer and thus a fixed threshold results in too aggressive pruning for initial layers. 
Lastly, we find that the per-token pruning strategy performs comparably to the per-layer strategy, but with the additional benefit of pruning a constant number of activations for each token. For this reason, we only consider top-K pruning in the remainder of this paper.

\subsection{Strategies for Dynamic Sparsity}
We now consider several dynamic weight sparsification schemes that exploit dynamic magnitude-based activation pruning. Figure~\ref{fig:sparsification_diagrams}a-c shows the different schemes for GLU LLMs described in the remainder of this section.

\paragraph{GLU pruning (Figure~\ref{fig:sparsification_diagrams}a)}
When values of $\GLU(\rvx)$ are pruned, the corresponding columns of $\Wd$ can also be pruned as they will be multiplied with a zero:
\begin{align}
    \aWd_{:,i} = \begin{cases}
        \boldsymbol{0}\ \ \ \ \ \ \ \ \ \ \text{if }i \notin \text{topK}\left(|\GLU(\rvx)|\right)\\
        \Wd_{:,i} \ \ \ \ \ \text{o.w.}
    \end{cases}.
\end{align}
Note that, instead of top-K pruning, other thresholding methods (such as those presented in Section~\ref{sec:magnitude_based_pruning}) can be used instead.

This strategy can sparsify only one of the three large MLP matrices, reaching at most $33 \%$ MLP sparsity.

\paragraph{Gate pruning (Figure~\ref{fig:sparsification_diagrams}b)}
Assuming that most of the small entries are determined by the non-linearity $\sigma$, one could first perform a dense matrix multiplication to determine $\sigma(\Wg\rvx)$, then sparsify both $\Wu$ and $\Wd$ based on the small (partial) activations:
\begin{align}
    \aWu_{:,i},\aWd_i, = \begin{cases}
        \boldsymbol{0},\boldsymbol{0}\ \ \ \  \text{if } i \notin \text{topK}\left(|\sigma(\Wg\rvx)|\right)\\
        \Wu_{:,i},\Wd_i \ \ \text{o.w.}
    \end{cases}.
\end{align}

Similarly, we can define  \textit{Up Pruning} as the equivalent strategy in which the activations $\Wu\rvx$ are used to prune $\Wg$ and $\Wd$, respectively.
In contrast to GLU pruning, Gate (and Up) pruning can prune up to $66 \%$ of the MLP weights. However, this pruning strategy may be sub-optimal since the network parameters are pruned based on partial GLU activations.

\paragraph{Predictive GLU pruning (Figure~\ref{fig:sparsification_diagrams}c)} 
The same predictive pruning strategy introduced by DejaVu can be extended to GLU LLMs by training a small predictor to identify low-magnitude GLU activations:
\begin{align}
    \aWu_{:,i},\!\aWg_{:,i},\!\aWd_{i}\!\! =\!\! \begin{cases}
        \boldsymbol{0},\!\boldsymbol{0},\!\boldsymbol{0}\ \ \ \ \text{if }i\notin\topk(\pred(\rvx))\\
        \Wu_{:,i},\Wg_{:,i},\Wd_{i} \ \ \text{o.w.}
    \end{cases}.
\end{align}
Here $\pred(\rvx)$ indicates the logits of a predictor trained to identify the smallest activations using cross-entropy loss.

Predictive GLU pruning methods can sparsify all MLP weight matrices, potentially resulting in higher sparsity levels at the same model accuracy. 
However, as we will show next, it is extremely challenging to train predictors for GLU pruning.


\subsection{The challenge of predicting GLU activations}
\begin{figure}[tb]
    \centering
    MMLU (5-shots)
    \includegraphics[width=0.9\columnwidth, trim={0 0 0 0.9cm}, clip]{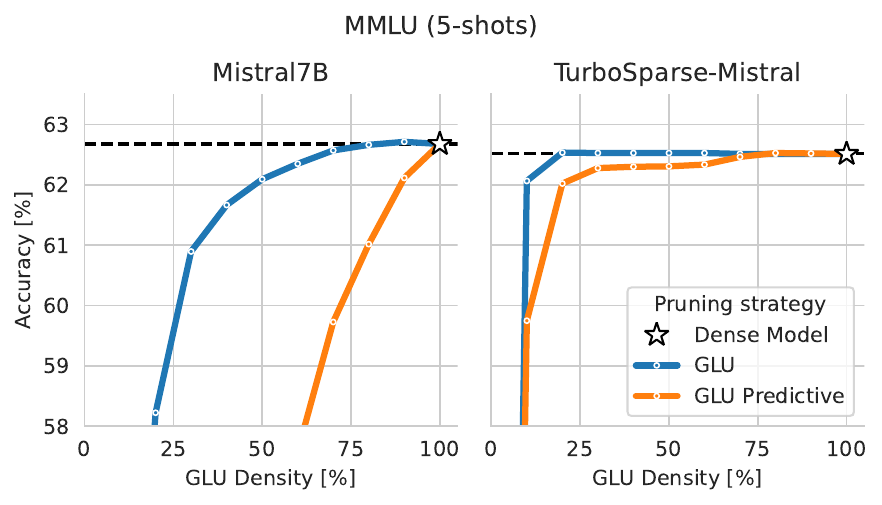}
    \vspace{-.2cm}
    \caption{Comparison of the effectiveness of GLU sparsification methods on Mistral-7B (left) and its ReLU-fied counterpart (right). The predictor training procedure, architecture, and training data is equivalent across the two LLMs.} 
    \label{fig:difficult_pruning}
\end{figure}

As demonstrated in Figure~\ref{fig:difficult_pruning} (blue line),
GLU pruning with $\topk$ thresholding allows us to prune large portions of activations with a minor impact on downstream performance.
On SwiGLU networks, such as Mistral-7B, it is possible to prune up to 50\% of the activations before observing a 1\% decrease in MMLU performance. 
However, the corresponding ReLU-fied network, TurboSparse-Mistral, is much more sparsifiable, and GLU pruning can reach more than 90\% activation sparsity before observing the same performance deterioration.
This showcases that SwiGLU activations are inherently more sensitive to pruning.
 
Without any predictive mechanism,
the effectiveness of GLU pruning is further hindered by the fact that only one of three MLP matrices is sparse (Figure~\ref{fig:sparsification_diagrams}a). 
We therefore studied the effect of predictive GLU pruning, which employs predictors trained with cross-entropy loss to identify small GLU activations.
Consistently with \citet{liu2023deja}, after training we prune the GLU activations corresponding to the smallest outputs of the predictor $pred(\rvx)$ on a per-token basis.
The results are visualized in Figure~\ref{fig:difficult_pruning} (orange line), which compares the effectiveness of predictive pruning on SwiGLU and ReLU-fied counterparts for the same set of predictors and training procedure.

MMLU accuracy at 50\% activation sparsity drops by almost 10\% on Mistral-7B, while the performance drop is negligible on the ReLU-fied counterpart.
The wider gap between GLU pruning and GLU predictive pruning hints that predicting activation magnitude is a much harder task in SwiGLU networks.
We hypothesize that this phenomenon is due to the interplay of various aspects. First, since SwiGLU activations are less sparse, predictor mistakes are more likely to prune relevant activations. Secondly, the activation pattern in GLU MLPs is less predictable than the one in the ReLU-fied counterparts. 
We conjecture that this is because predicting ReLU zeros is essentially sign prediction for a linear projection, a relatively easy task, whereas predicting magnitude is a more challenging task due to the interplay between the gating components.

Due to the inherent limitations of GLU pruning and the challenges of predicting complex SwiGLU activation patterns with small predictors, we consider a simple alternative approach that aims to sparsify all three MLP matrices without requiring any additional predictor architecture.


\section{Dynamic Input Pruning}
\label{sec:dynamic_input_pruning}

In this section we introduce the Dynamic Input Pruning (\textbf{DIP}) strategy, which addresses the limitations of existing MLP dynamic pruning strategies on SwiGLU networks.
The DIP method, visualized in Figure~\ref{fig:sparsification_diagrams}d, hinges on removing small activations in both the \emph{input} of each MLP block and the \emph{intermediate} GLU block using per-token $\topk$ thresholding, and thus requires no separate predictors. 
DIP is defined as:
\begin{align}
    \aWu_{:,i},\aWg_{:,i} &=   \begin{cases}
        \boldsymbol{0},\boldsymbol{0}\ \ \ \ \ \ \ \ \ \ \ \ \ \text{if } i \notin \text{topK}\left(|\rvx|\right)\\
        \Wu_{:,i},\Wg_{:,i}\ \ \text{o.w.}
    \end{cases}\\
    \aWd_{:,i} &=   \begin{cases}
        \boldsymbol{0}\ \ \ \ \ \ \ \ \text{if } i \notin \text{topK}\left(|\aGLU(\rvx)|\right)\\
        \Wd_{:,i}\ \ \text{o.w.}
    \end{cases}.
\end{align}

$\aGLU(\rvx)$ corresponds to the GLU activations obtained using $\aWu$ and $\aWg$ instead of $\Wu$ and $\Wg$. Note that we can use different levels of sparsity for the up, gate and down layers. The optimal proportion of active units is empirically determined as a function of the overall target MLP sparsity (see Appendix~\ref{app:tune_up_down_sparsity} for details).

When compared to Predictive GLU pruning, Dynamic Input Pruning trades off the error caused by the predictor for approximation error introduced by the use of an approximate gating layer.

\paragraph{Light-weight LoRA fine-tuning}
To compensate for the approximation error caused by the subsequent pruning of multiple MLP activations, similarly to contemporary literature \citep{ma2023llm, sun2023simple, lee2024cats}, we add fine-tuned LoRA adapters to the sparsified up, gate and down MLP matrices.
To ensure that the LoRA adapters do not lead to any additional overhead, each adapter is applied before column selection:
\begin{align}
    \underbrace{\mW_{:,i}}_{\text{original column}} \to \underbrace{\left(\mW + \mA\mB\right)_{:,i}}_{\text{LoRA adapted column}}, 
\end{align}
in which $\mA$ and $\mB$ refer to low-rank learnable matrices. This allows us to fuse the adapters to the original matrices after fine-tuning to avoid any memory overhead.

\section{Improved Caching in Dynamic Sparsity}
\label{sec:method_cache}

An effective dynamic sparsity method is not enough to guarantee good LLM inference latency on a memory-constrained device. The number of Flash reads, which bottlenecks the overall throughput, largely depends on the cache eviction policy employed to dynamically load and unload weights in DRAM during inference.
In this section we discuss existing cache eviction policies and present a novel cache-aware masking strategy for dynamic sparsity. Our approach allows us to reach better accuracy-latency trade-offs with respect to solely regulating the sparsity level.

\begin{figure}[t]
    \centering
    \includegraphics[width=0.9\columnwidth]{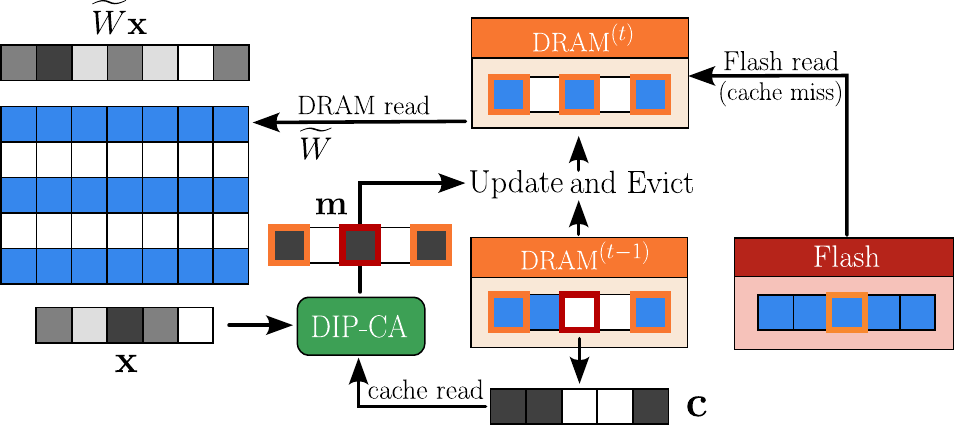}
    \caption{Diagram for the Cache-aware Dynamic Input Pruning method applied to an individual linear layer. DIP-CA uses cache state and input activations to output a sparsity mask. Flash reads are performed to load necessary weights missing from cache (marked in red), and unused weights are evicted if the DRAM is full. The sparse weight matrix is then used to process the current activations.} 
    \label{fig:hw_diagram}
\end{figure}

\subsection{Cache eviction policies}
A cache management system can be employed to optimize the decision of which parameters to evict, when the cache is full and we want to allocate new weights. 
Commonly used cache eviction policies include Least Recently Used (LRU) and Least Frequently Used (LFU). In LRU, the item that was last used farthest in the past is evicted. In LFU, a counter is kept (for the current session or across sessions) to track usage frequency and evict the most rarely used item.

An optimal cache eviction policy can be implemented if future DRAM reads were known in advance. Belady \yrcite{belady1966study} describes a greedy oracle algorithm to maximize the cache hit-rate over a known, fixed sequence of memory reads. Based on this optimal algorithm, previous work \cite{lykouris2021competitive, liu2022space, sadek2024algorithms} explores methods to use estimators \cite{jain2016back, shah2022effective} or learnable ML models based on cache history to predict the future cache usage. While these methods outperforms simpler caching eviction policies, they require an additional training step, and their performance is upper-bounded by the optimal Belady caching algorithm.

Caching methods for dynamic sparsity have also been employed, with the eviction policy used to determine which parameters should be kept in DRAM to increase the cache hit rate and therefore minimize latency. 
For example, \citet{song2023powerinfer} use a static cache by preloading the most active neurons, \citet{alizadeh2023llm} introduce an LRU eviction policy to prune unused weights, and \citet{xue2024powerinfer} propose neuron bundling over multiple weight matrices.

\subsection{Cache-aware neuron sparsity}

Differently from other caching applications, the current choice of sparsity mask is not a hard constraint. We argue that suboptimal masks can be preferred to increase the system throughput while sacrificing some model accuracy. We thus propose a \textit{cache-aware} sparsification method to strike a better trade-off between accuracy and latency. This is implemented as a re-weighting mechanism before selecting active neurons, which introduces a penalty multiplier term on the neurons activation magnitude:
\begin{align}
    \st = \cfrac{\left| \xt \right| \, \left(\ct + \gamma (1 - \ct) \,  \right)}{\left\Vert \xt \right\Vert_{\infty}}, \label{eq:deweighting} 
\end{align} 
where $\xt$ refers to the current input activations, $\ct$ is a binary mask representing which neurons are currently loaded in cache, and $\gamma$ is a hyper-parameter to penalize the selection of non-cached neurons. The resulting scores $\st$ are used instead of the original magnitude for dynamic pruning as discussed in Section~\ref{sec:dynamic_input_pruning}. The normalization term in the denominator ensures that the re-weighting mechanisms is not sensitive to changes in the dynamic range of the activations over different tokens.

Our implementation choice is motivated by the observation that for each layer and token, few neurons fire very strongly, while activations for most neurons are orders of magnitude lower (see Figure~\ref{fig:cache_gamma}, left), and have less impact on the layer outputs. 
The penalty term is thus set high enough to allow re-ranking of the activations in the intermediate regime, but low enough to not impact the strongest activations even if not in cache. We empirically find that a simple LFU eviction policy with cache-aware masking results in better latency-accuracy trade-offs even compared against the oracle algorithm from Belady \yrcite{belady1966study} without re-weighting (see Section \ref{sec:cache_res}).

We name \textbf{DIP-CA} the Cache-Aware variant of Dynamic Input Pruning, and show it outperforms DIP in terms of throughput at all perplexity levels. Figure~\ref{fig:hw_diagram} visualizes the DIP-CA procedure, which is described in detail in Algorithm \ref{alg:cache-aware} for a single linear layer.

\begin{algorithm}[t]
\caption{Dynamic Input Pruning with Cache-Aware re-weighting for a single linear layer.}
\label{alg:cache-aware}
\begin{minipage}{0.9\columnwidth}
\textbf{function} \, $\operatorname{DIP-CA}\left(\xt, \ct, \W, \gamma, k\right)$:
\begin{algorithmic}\doublespacing
    \State $\st = \cfrac{\left| \xt \right| \, \left(\ct + \gamma (1 - \ct) \,  \right)}{\left\Vert \xt \right\Vert_{\infty}}  $
    \!\! \Comment Eq.\ref{eq:deweighting}
    \State $\mt = \topk\left( \st, k \right)$
    \State $\ctt = \update\left( \ct, \mt\right)$
    \State $\aW^{(t)} = \mW \odot {\mt}$  
    \State \Return \; $\aW^{(t)}, \ctt$
\end{algorithmic}
\end{minipage}
\end{algorithm}

\section{Evaluation}
\label{sec:evaluation}

\begin{table*}[h!]
    \centering
    \caption{Experimental results for dynamic sparsity methods at 50\% MLP sparsity. $^*$The MLP density estimation ignores the overhead introduced by the predictors or additional data-structures. GLU pruning is not included since at most 33\% sparsity can be achieved.} 
    \label{tab:exp1}
    \vskip 0.1in
    
    \begin{small}
    \begin{tabular}{l|cccc|cccc}  
        \toprule
        & \multicolumn{4}{c|}{WikiText-2 (Perplexity $\downarrow$) } & \multicolumn{4}{c}{MMLU (5-shot accuracy $\uparrow$) } \\
        \cmidrule(r){2-9}
         & Phi3Med & Phi3Mini & Llama8B & Mistral7B & Phi3Med & Phi3Mini & Llama8B & Mistral7B \\
        \midrule
        Dense & 4.29 & 6.01 & 6.14 & 5.25 & 78.14 & 70.62 & 65.30 & 62.68 \\
        GLU Pruning (oracle) & 4.45 & 6.10 & 6.30 & 5.29 & 77.98 & 70.24 & 64.77 & 62.09 \\
        \midrule
        SparseGPT (unstructured)$^*$ & 5.67 & 7.44 & 7.71 & 5.81 & 73.97 & 62.15 & 57.72 & 58.16 \\
        SparseGPT (2:4)$^*$ & 6.87 & 9.84 & 10.05 & 7.31 & 68.07 & 53.39 & 43.69 & 51.15 \\
        SparseGPT (4:8)$^*$ & 6.33 & 8.43 & 8.64 & 6.41 & 70.34 & 57.98 & 50.78 & 54.64 \\
        \midrule
        Gate Pruning & 11.28 & 14.83 & 29.26 & 96.37 & 66.08  & 47.90 & 27.23 & 40.39 \\
        Up Pruning & 7.41 & 10.94 & 13.93 & 7.13 & 70.68 & 56.82 & 46.34 & 50.09 \\
        DejaVu$^*$ & 6.15 & 8.76 & 9.43 & 6.24 & 68.99 & 54.86 & 49.02 & 54.34 \\
        CATS & 8.34 & 10.83 & 28.24 & 60.48 & 71.08 & 55.64 & 27.34 & 38.96 \\
        CATS+LoRA & 5.54 & 7.44 & 16.53 & 9.18 & 72.23 & 60.16 & 30.52 & 44.84 \\
        \textbf{DIP} & 5.52 & 7.04 & 7.38 & 5.61 & 75.54 & 66.51 & 60.38 & \underline{59.66} \\
        \textbf{DIP+LoRA} & \underline{5.01} & \underline{6.79} & \underline{7.26} & \underline{5.57} & \underline{75.89} & \underline{66.57} & \underline{60.94} & 59.58 \\
        \bottomrule
    \end{tabular}
    \end{small}
\end{table*}

\subsection{Experimental setup}
\label{sec:exp_setup}

\paragraph{A three-way trade-off in KPIs} 
We observe how methods for efficient LLM inference must be evaluated based on three different metrics: 1) model  \textit{accuracy}, 2) \textit{memory footprint} and 3) inference \textit{throughput}. Depending on the target application, a different trade-off between these metrics can be considered optimal. To simplify the experimental analysis, we evaluate at most two metrics at a time, while ignoring or fixing the remaining ones. We first compare model performance at a fixed sparsity rate, that is at the same constraint in terms of memory footprint. The model accuracy is reported in terms of token perplexity or 5-shot accuracy on downstream tasks. We then evaluate the throughput achievable at different operating points in terms of both memory footprint and loss in model accuracy. We include in Appendix ablations on hardware specifications to show how different methods perform in various use-cases.

\paragraph{HW Simulator} 
In order to accelerate the evaluation procedure,
we implement a software tool to simulate the data transfers between Flash, DRAM and processing units, and then compute the latency incurred in these processes. We also implement simulators for different cache eviction policies on DRAM. This allows us to efficiently estimate the impact of different dynamic sparsity solutions on system latency, without having to implement or modify a full system on-device. As inference in the token generation scenario is bound by DRAM and Flash accesses \cite{jhang2021challenges, kwon2023efficient, gholami2024ai}, we do not simulate NPU inference times, which can be largely parallelized or are negligible when compared to the more expensive memory operations. In Appendix \ref{app:hw_sim} we include a detailed description of the simulator, including motivation for the selected parameters.

Unless otherwise stated, results throughout the paper are based on simulations for Apple A18 with DRAM I/O speed of $60$ GB/s, and Flash read speed of $1$ GB/s. We also conduct an ablation study on different choices for DRAM size and Flash read speed. Results are reported in Appendix \ref{app:ablation_flash}.

\paragraph{Methods and Models} 
We evaluate the proposed approach against the Gate Pruning, Up Pruning, and GLU Pruning methods presented in Section \ref{sec:related_work}. We further consider CATS \cite{lee2024cats} and DejaVu \cite{liu2023deja}: two recent dynamic sparsity solutions. Following the original approach, each predictor consist of a neural network of 1000 hidden units, which is trained for at most 20 epochs on activations collected from a calibration set using cross entropy loss. For SwiGLU LLMs, the binary targets are generated by selecting the 10\% largest activations for each token.
We further compare Dynamic Input Pruning with unstructured and semi-structured (2:4, 4:8) SparseGPT pruning \cite{frantar2023sparsegpt} and quantization methods GPTQ \cite{frantar2022gptq} and GPTVQ \cite{van2024gptvq}. 

In DIP, we empirically calibrate the proportion of active units for a target MLP sparsity as described in Appendix~\ref{app:tune_up_down_sparsity}.
For DIP-CA, we tune the cache-aware re-weighting hyper-parameter on WikiText-2 validation set and fix it to $\gamma=0.2$ for all experiments. We include an ablation showing the impact of changes in $\gamma$. Following previous work \citep{hu2021lora, lee2024cats}, we train LoRA adapters with rank 32 for a total of 1000 iterations using a knowledge distillation loss to match the original (dense) model logits. The training procedure is consistent for DIP and CATS models. The adapters are applied to all the \textit{up}, \textit{down} and \textit{gate} matrices for DIP, while only \textit{up} and \textit{down} are adapted for CATS.

We evaluate each method on four recent SwiGLU LLMs with different dimensionalities: Phi-3-Medium and Phi-3-Mini \cite{abdin2024phi}, Llama 3 8B \cite{dubey2024llama}, Mistral 7B \cite{jiang2023mistral}.

\paragraph{Datasets}
We use the SlimPajama \cite{cerebras2023slimpajama} dataset as calibration datasets to train predictors and fine-tune LoRA adapters. We evaluate all methods on token perplexity for language modeling on WikiText-2 test set with sequence length $2048$, and on 5-shot accuracy for downstream tasks on the MMLU benchmark \cite{hendrycks2020measuring} using the LM Evaluation Harness \cite{eval-harness}.

\begin{figure}[t!]
    \centering
    \includegraphics[width=\columnwidth, trim={0 0 0 0.9cm}, clip]{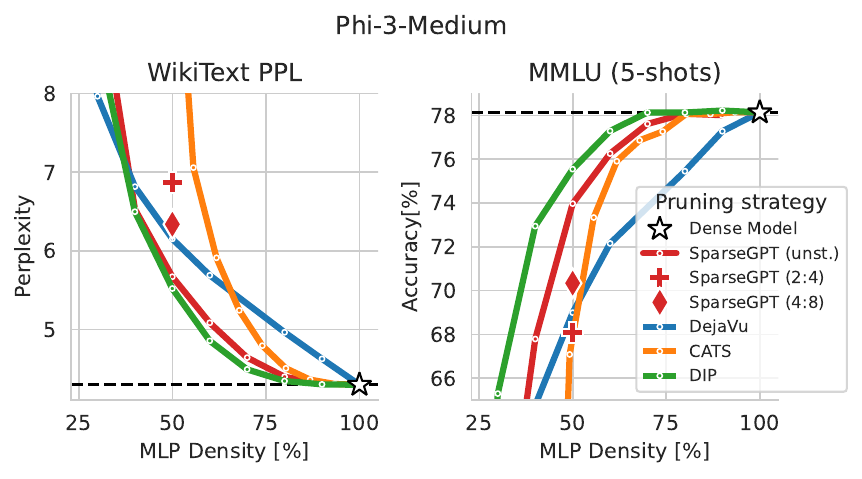}
    \vspace{-0.2cm}
    \caption{Pareto-curves for WikiText-2 perplexity (left) and 5-shot MMLU accuracy (right) vs MLP density on Phi-3-Medium. The density estimation ignores overhead in DejaVu and SparseGPT.} 
    \label{fig:ppl_phi_medium}
\end{figure}

\subsection{Results}

\paragraph{Memory footprint} We firstly compare DIP against recent static and dynamic pruning methods. In Figure~\ref{fig:ppl_phi_medium} we show the trade-off between model performance and MLP density obtained at varying sparsity levels. Note that certain methods require additional parameters to achieve the sparsification, but we do not account for this in the visualizations, as the resulting overhead depends on the bit-width choice for quantization. DejaVu \cite{liu2023deja} requires sparsity predictors which amount up to 15\% of the number of parameters for the dense MLP. For SparseGPT \cite{frantar2023sparsegpt} unstructured and structured pruning, at least 1 extra bit of information for each weight is needed to indicate whether the weight is pruned \cite{kuzmin2024pruning}. The memory overhead in this case would be 6.25\% and 25\% for 16-bit and 4-bit quantized models, respectively. We observe how DIP outperforms both static and dynamic pruning methods at all sparsity levels, even excluding this overhead. Noticeably, we can run Phi-3-Medium in 70\% of its full memory footprint, at the cost of only 0.20 perplexity and 0.01\% accuracy with respect to the dense model. In comparison, CATS \cite{lee2024cats} incurs in a 0.50 increase in perplexity and 0.88\% decrease in accuracy, while for DejaVu \cite{liu2023deja} that is 1.03 higher perplexity and 4.34\% lower accuracy. We include in Appendix \ref{app:additional_results} additional results for Phi-3-Mini, Llama-v3-8B, and Mistral-7B.

In Table \ref{tab:exp1} we report more extensive numerical results on four modern LLMs. We consider 50\% MLP sparsity for each method, and report the perplexity on WikiText-2 as well as the 5-shot accuracy on the MMLU benchmark. Results for different operating points can be found in Table~\ref{tab:exp1_40} and Table~\ref{tab:exp1_60} provided in Appendix \ref{app:additional_results}, which also includes evaluations on a broader range of tasks.

Static pruning methods \cite{frantar2023sparsegpt}, particularly in the semi-structured setup, show a clear gap in accuracy with respect to the dense model.
Gate Pruning and Up Pruning are simple baselines introduced in Section~\ref{sec:dynamic_sparsity}. Pruning activations based on the output of the \textit{gate} activations $\sigma(\Wg\rvx)$  (Gate Pruning) severely affects performance on LLama-v3-8B and Mistral-7B. Noticeably, pruning gate layers based on the output of the up component $\Wu\rvx$ (Up Pruning) instead gives better performance across all tested architectures. Still, both approaches are far from the ideal results achievable with an oracle for GLU Pruning. This oracle shows that, with perfect predictions, model performance would be marginally worse than the dense baseline. 

CATS \cite{lee2024cats} is an instance of per-layer threshold pruning on the gate activations. Comparably with Gate Pruning, the method struggles with the Llama-v3-8B and Mistral-7B models. 
We expect that modifying the CATS approach to use up layer activation might improve its effectiveness. Introducing LoRA adapters over CATS is necessary to boost performance. In contrast, our proposed DIP method without adapters outperforms CATS+LoRA as well as previous baselines on all models and evaluation scenarios.
DIP+LoRA further improves overall results, with more pronounced benefits on perplexity scores.

\begin{table}[t]
    \centering
    \caption{Comparison of throughput for dynamic sparsity methods. We report the highest throughput achieved at a 0.2 and 0.5 increase in perplexity on WikiText-2 over the dense model. The available DRAM capacity is set to fit approximately 50\% of each dense model. All models are 4-bit quantized.} 
    \label{tab:exp2}
    \vskip 0.1in
    
    \begin{small}
    \setlength{\tabcolsep}{4.5pt}
    \begin{tabularx}{\columnwidth}{l|cccc}  
        \toprule
         & Phi3Med & Phi3Mini & Llama8B & Mistral7B \\
        \midrule
        Model size & 7.4 GB & 2.4 GB & 4.3 GB & 3.9 GB \\
        DRAM size & 4.0 GB & 1.5 GB & 2.5 GB & 2.0 GB \\
        \toprule
        & \multicolumn{4}{c}{Throughput [tok/s] ($\uparrow$) } \\
        \midrule
        Dense           & 0.29 & 1.15 & 0.57 & 0.54 \\
        \toprule
        & \multicolumn{4}{c}{Throughput [tok/s] @ + 0.2 PPL ($\uparrow$) } \\
        \midrule
        GLU Pruning     & 0.41 & 1.59 & \underline{0.78} & 0.80 \\
        Up Pruning      & 0.43 & 1.69 & 0.76 & 1.03 \\
        CATS            & 0.40 & 1.54 & 0.73 & 0.69 \\
        \textbf{DIP}     & 0.43 & 1.69 & 0.73 & 0.98  \\
        \textbf{DIP-CA} & \underline{0.45} & \underline{1.77} & \underline{0.78} & \underline{1.08}  \\
        \toprule
        & \multicolumn{4}{c}{Throughput [tok/s] @ + 0.5 PPL ($\uparrow$) } \\
        \midrule
        GLU Pruning     & 0.45 & 1.63 & 0.84 & 0.87 \\
        Up Pruning      & 0.52 & 1.98 & 0.87 & 1.32 \\
        CATS            & 0.47 & 1.84 & 0.83 & 0.77 \\
        \textbf{DIP}    & 0.50 & 1.98 & 0.82 & 1.23  \\
        \textbf{DIP-CA} & \underline{0.56} & \underline{2.09} & \underline{0.89} & \underline{1.39}  \\
        \bottomrule
    \end{tabularx}
    \end{small}
\end{table}

\paragraph{Model throughput} In Table~\ref{tab:exp2} we investigate what throughput can be achieved with different sparsity methods and models. For this analysis, all models are quantized to INT4 using GPTQ \cite{frantar2022gptq}, and all methods rely on an LFU cache allocated as explained in Section~\ref{sec:exp_setup}. We consider a use-case where approximately half of the dense model can fit in DRAM and include the selected DRAM sizes compared against the model sizes in the first part of the table. To numerically evaluate the trade-off between model accuracy and efficiency, we report the highest throughput achieved by each method under fixed operating points in perplexity loss. 
We first notice that GLU Pruning underperforms Up Pruning, as GLU Pruning needs to load the full gate and up layers to compute GLU activations, which do not fit entirely in DRAM. CATS \cite{lee2024cats} falls short from the previous methods, possibly because of the sub-optimality of Gate Pruning described in Table \ref{tab:exp1}. By itself, DIP reaches similar or marginally worse throughput than Up Pruning.
Finally, we assess the benefits of applying Cache-Aware masking to DIP. For all models and perplexity operating points, DIP-CA outperforms previous methods. Noticeably, DIP-CA enables running Phi-3-Medium with 46\% lower DRAM footprint and 40\%, 55\% and 93\% faster throughput at a perplexity loss of 0.1, 0.2 and 0.5, respectively. Similar improvements are observed for Phi-3-Mini, LLama-v3-8B and Mistral-7B.

\subsection{Comparison and compatibility with quantization and static pruning}

\begin{figure}[t]
    \centering
    \includegraphics[width=0.9\columnwidth, trim={0 0.2cm 0 0}, clip]{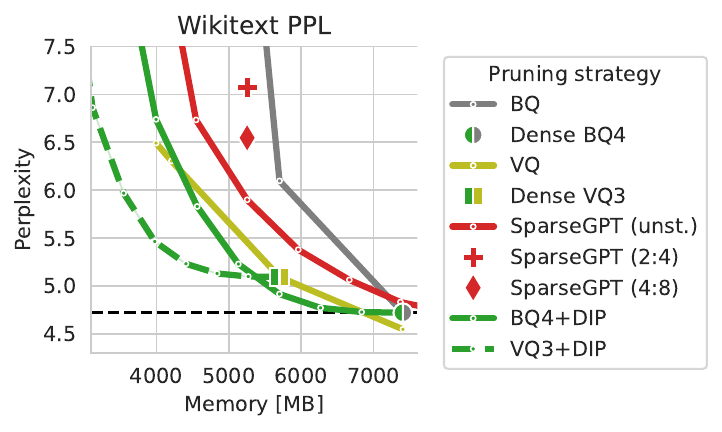}
    \caption{Comparison against quantization and pruning, integration of DIP with 4 bit BQ and 3 bit VQ models on Phi-3-Medium. } 
    \label{fig:quant_pruning}
\end{figure}

Quantization is an alternative approach to improve model efficiency. We compare DIP with Blockwise Quantization (BQ) \cite{frantar2022gptq} and Vector Quantization (VQ) \cite{van2024gptvq} at 2, 3 and 4 bits, as well as static pruning with SparseGPT \cite{frantar2023sparsegpt}. 
Note that in these results we do account for 1 bit of overhead per weight introduced by the pruning mask.
In Figure \ref{fig:quant_pruning} we report results on the comparison and combination of DIP with quantization. 
DIP applied to 4 bit BQ models compares favorably against both pruning and quantization approaches, reaching 15\% lower perplexity than SparseGPT given 4GB of available DRAM.
At higher quantization levels, BQ4+DIP outperforms more aggressive BQ quantization, which suggests that increased dynamic sparsity is preferred over decreasing quantization bit-width.
VQ3+DIP improves over BQ4+DIP at higher sparsity, similarly outperforming 2-bit VQ, and showing that DIP can be considered as complementary to quantization methods to enhance model efficiency.

\subsection{Cache eviction policies}
\label{sec:cache_res}

\begin{figure}
    \centering
    \begin{minipage}{0.46\columnwidth}
        \includegraphics[width=\textwidth]{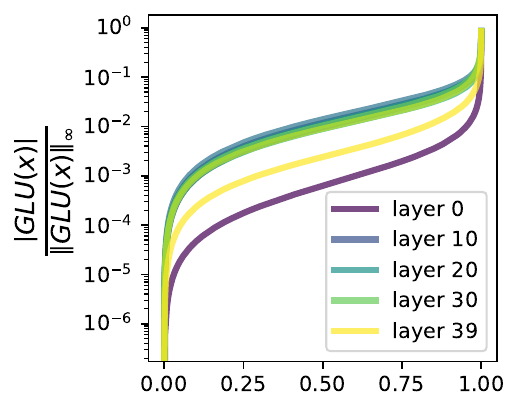}
    \end{minipage}
    \begin{minipage}{0.50\columnwidth}
        \includegraphics[width=\textwidth]{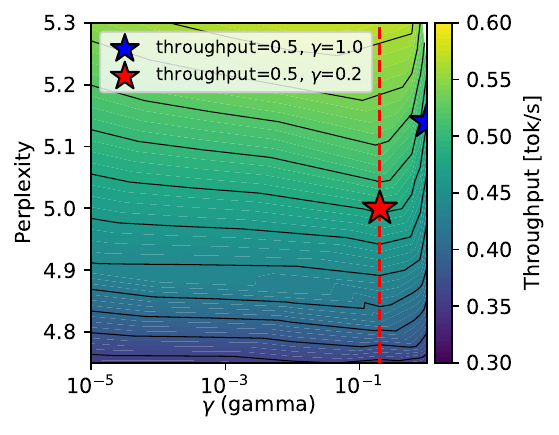}
    \end{minipage}
    \caption{(Left) Distribution of normalized GLU activations for Phi-3-Medium layers. 
    (Right) Effects of changing the DIP-CA $\gamma$ hyper-parameter on throughput at different levels of perplexity.}
    \label{fig:cache_gamma}
\end{figure}

\paragraph{Cache-aware masking}
By analyzing the distribution of GLU activations in Phi-3-Medium (see Figure~\ref{fig:cache_gamma}, left), we observe that a small percentage of activations are orders of magnitude larger than the rest, and will have a bigger impact in determining the layer output. On the other hand, activations between the $30^{th}$ and $80^{th}$ percentiles are within one order of magnitude, which means they will have a relatively similar contribution to the linear layer output. DIP-CA exploits this behavior by re-weighting the scores for activations depending on whether they are currently available in cache. This effectively re-orders the activation scores before the TopK selection. In Figure \ref{fig:cache_gamma} (right), we show the effect of changing the $\gamma$ hyper-parameter in DIP-CA. For all perplexity levels, $\gamma$ ranges between 0.1 and 0.3 result in optimal throughput. We include an example by comparing two models achieving the same throughput. The DIP instance with $\gamma = 1$ (in blue) has a perplexity of 5.14, while the DIP-CA counterpart with $\gamma = 0.2$ reaches lower perplexity of 5.0.
Since the activation distributions are comparable across layers (with first and last layers having slightly longer tails), we use a fixed value of $\gamma$ for all layers.

\begin{figure}[t]
    \centering
    \begin{minipage}{0.9\columnwidth}
        \includegraphics[width=\textwidth, trim={0 0.1cm 0 0}, clip]{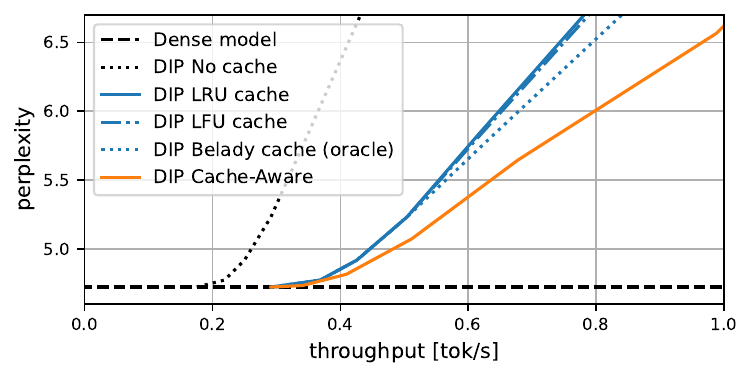}
    \end{minipage}
    \caption{Comparison between cache eviction policies and cache-aware masking. LFU with DIP-CA outperforms all policies including Belady's oracle throughput at a given perplexity threshold.}
    \label{fig:cache_main}
\end{figure}

\paragraph{Comparison with cache eviction policy}
We compare in Figure ~\ref{fig:cache_main} different caching policies to handle allocation and eviction of weights in DRAM. 
Without 
DRAM caching, inference inference throughput is severly degraded, as the MLP weights need to be loaded from Flash for each token. 
LFU and LRU policies regulate the eviction of parameters once the DRAM cache is full, by considering how often or recently a neuron in the cache has been used.
Results are very similar, with LFU being marginally better. We then observe how even the oracle Belady cache eviction policy does not yield much improvement over practical policies.
The throughput for all caching eviction policies overlap under 5.25 perplexity, corresponding to the sparsity level under which the parameter size for the active neurons becomes larger than the DRAM size allocated for caching. In this case, the cache will always contain current neurons, or part of them. 
In contrast, DIP-CA with a simple LFU policies largely outperforms DIP with all cache eviction policies, even including the optimal Belady algorithm. The improvement also holds at lower sparsity regimes, where part of the active neurons are directly loaded to the processing unit, as there is no remaining space to allocate them in cache.

\section{Conclusion}
\label{sec:conclusion}

This paper addresses the challenge of efficient LLM inference on DRAM constrained mobile devices.
Existing dynamic sparsity approaches are not effective for modern 
LLMs which use SwiGLU instead of ReLU, and hence do not exhibit natural sparsity. 
Instead, we introduce a novel dynamic input pruning (DIP) solution which removes the need for predictors and requires only minimal fine-tuning. DIP can be further aided by 
a lightweight LoRA adapter to recover part of the accuracy lost during sparsification. 
Finally, we introduced a cache-ware variant (DIP-CA),
which significantly enhances model throughput and reduces memory footprint. 
Specifically, we achieved a 46\% reduction in memory usage and a 40\% increase in throughput with respect to dense 
Phi-3-Medium,
with a minimal impact on perplexity. 
Underscoring these results, we show extensive comparisons against pruning and quantization baselines, as well as ablation studies considering different hardware specifications.

Future work will focus on further refining our sparsity techniques and exploring their applicability to a broader range of models and hardware configurations, including on-device simulation of model performance.


\bibliography{main}
\bibliographystyle{mlsys2024}


\clearpage
\appendix

\twocolumn[{
    \centering
    \includegraphics[width=0.99\textwidth]{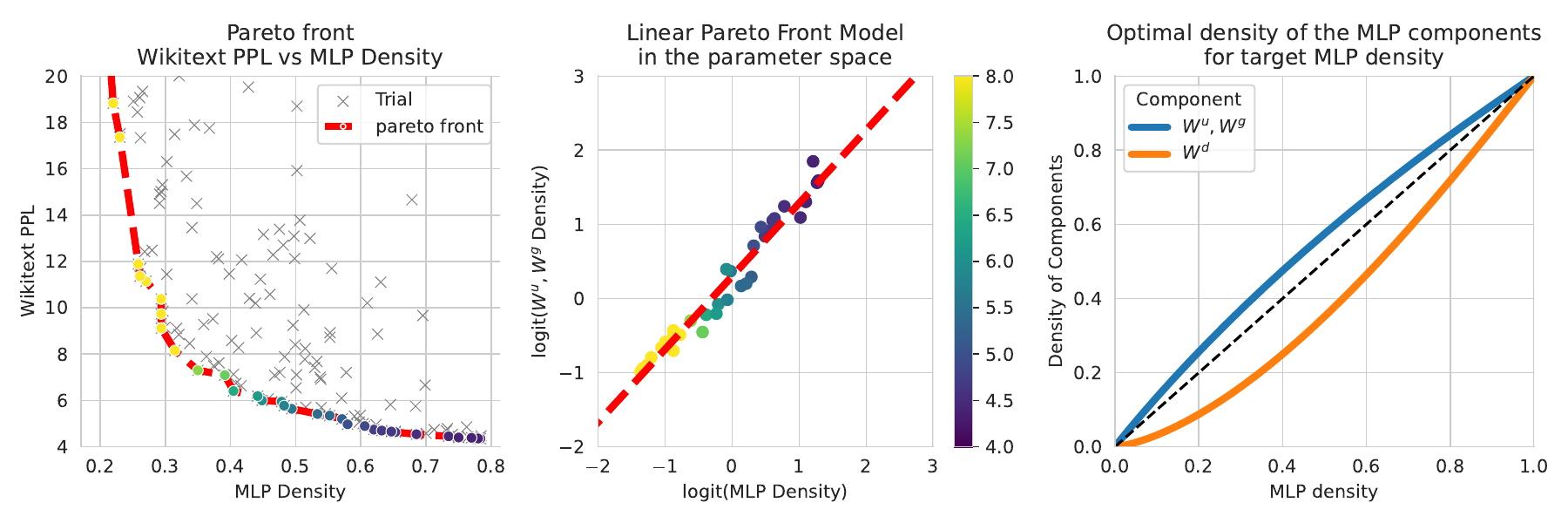}
    \captionof{figure}{Optimal allocation of weight density between $\aWd$ and $(\aWu,\aWg)$.}
    \label{fig:sup_down_alloc}
    
    \centering
    \includegraphics[width=0.6\textwidth]{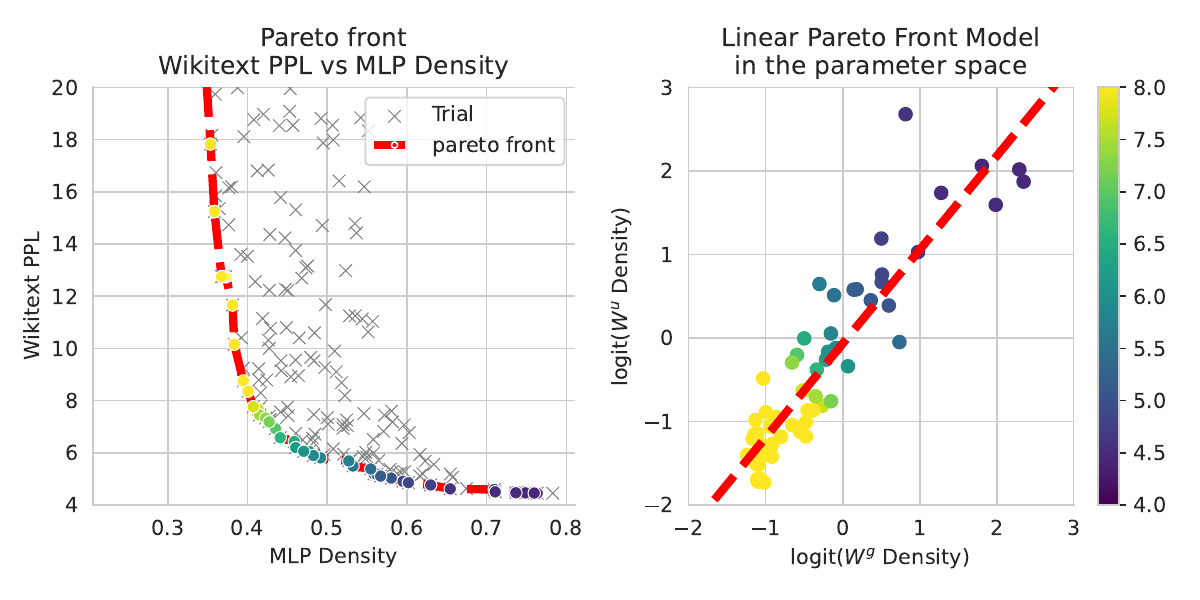}
    \captionof{figure}{Allocation of weight density between $\aWu$ and $\aWg$ matrices for $\aWd$ with fixed density of $50\%$ for \textit{phi-3-medium model} on the Wikitext dataset. 
    \\~\\}
    \label{fig:up_gate_alloc}
}]

\section{HW Simulator description}
\label{app:hw_sim}

The main parameters needed for simulation are therefore the DRAM capacity, and Flash and DRAM read/write speeds. We notice similar specifications in the processors for latest popular smartphone devices: Apple A18 \cite{enwiki_apple} and Snapdragon\textsuperscript{\textregistered} 8s Gen 3 SM8635 platform \cite{enwiki_snapdragon}. While latest UFS versions can reach up to 5.8 GB/s in bandwitdh \cite{enwiki_ufs}, the effective reading speeds might vary significantly depending on hardware interfaces and whether sequential or random reads are prevalent \cite{xue2024powerinfer}. The same applies for NVMe storage systems adopted by Apple devices \cite{labaran2023apple}. Unless otherwise stated, results throughout the paper are based on simulations for Apple A18 with DRAM I/O speed of $60$ GB/s, and Flash read speed of $1$ GB/s. 

A significant part of the available DRAM in smartphones is usually reserved by the OS or other applications \cite{daniel2023how}. We consider multiple values for available DRAM capacity to investigate how differently sized LLMs perform in memory-constrained scenarios. We include ablations with varying DRAM capacity (Table~\ref{tab:hardware_dram}) and Flash read speed (Table~\ref{tab:hardware_flash}) to assess whether the performance is consistent over several device specifications.

Consistently with \citet{xue2024powerinfer}, for cache simulation, we statically allocate to DRAM all the layers which do not undergo dynamic pruning, such as attention layers, embeddings, but also KV-cache and for certain methods, auxiliary modules like predictors. Since these layers are needed to process each token, pre-loading them in DRAM is an optimal choice \cite{xue2024powerinfer}. We then allocate the remaining DRAM capacity uniformly to all the MLP layers to enable the dynamic loading and caching of MLP weights. We did not find significant improvements when exploring non-uniform cache allocation.

\begin{table*}[h]
    \centering
    \caption{Experimental results for dynamic sparsity methods at 60\% MLP density.} 
    \label{tab:exp1_40}
    \vskip 0.1in
    
    \begin{small}
    \begin{tabular}{l|cccc|cccc}  
        \toprule
        & \multicolumn{4}{c|}{WikiText-2 (Perplexity $\downarrow$) } & \multicolumn{4}{c}{MMLU (5-shot accuracy $\uparrow$) } \\
        \cmidrule(r){2-9}
         & Phi3Med & Phi3Mini & Llama8B & Mistral7B & Phi3Med & Phi3Mini & Llama8B & Mistral7B \\
        \midrule
        Dense & 4.29 & 6.01 & 6.14 & 5.25 & 78.14 & 70.62 & 65.30 & 62.68 \\
        GLU Pruning (oracle) & 4.35 & 6.04 & 6.20 & 5.26 & 78.28 & 70.42 & 65.00 & 62.35 \\
        \midrule
        SparseGPT (unstructured) & 5.08 & 6.63 & 6.87 & 5.46 & 76.26 & 66.81 & 60.91 & 61.23 \\
        \midrule
        Gate Pruning & 6.36 & 7.97 & 10.22 & 8.67 & 74.85 & 62.14 & 51.62 & 55.53 \\
        Up Pruning & 5.68 & 7.51 & 8.50 & 5.74 & 75.54 & 64.67 & 60.29 & 59.91 \\
        DejaVu & 5.69 & 7.74 & 8.35 & 5.87 & 72.15 & 59.34 & 55.30 & 57.58\\
        CATS & 5.91 & 7.43 & 9.47 & 8.89 & 75.91 & 64.63 & 55.11 & 55.65 \\
        CATS+LoRA & 4.95 & 6.65 & 8.61 & 6.63 & 75.79 & 65.64 & 53.94 & 56.44 \\
        \textbf{DIP} & 4.85 & 6.4 & 6.66 & 5.39 & 77.29 & \underline{68.88} & \underline{63.17} & 61.19 \\
        \textbf{DIP+LoRA} & \underline{4.62} & \underline{6.35} & \underline{6.63} & \underline{5.38} & \underline{77.39} & 68.77 & 62.95 & \underline{61.50} \\
        \bottomrule
    \end{tabular}
    \end{small}
    \caption{Experimental results for dynamic sparsity methods at 40\% MLP density.} 
    \label{tab:exp1_60}
    \vskip 0.1in
    
    \begin{small}
    \begin{tabular}{l|cccc|cccc}  
        \toprule
        & \multicolumn{4}{c|}{WikiText-2 (Perplexity $\downarrow$) } & \multicolumn{4}{c}{MMLU (5-shot accuracy $\uparrow$) } \\
        \cmidrule(r){2-9}
         & Phi3Med & Phi3Mini & Llama8B & Mistral7B & Phi3Med & Phi3Mini & Llama8B & Mistral7B \\
        \midrule
        Dense & 4.29 & 6.01 & 6.14 & 5.25 & 78.14 & 70.62 & 65.30 & 62.68 \\
        GLU Pruning (oracle) & 4.64 & 6.24 & 6.52 & 5.35 & 77.74 & 69.23 & 64.28 & 61.67 \\
        \midrule
        SparseGPT (unstructured) & 6.53 & 9.51 & 9.68 & 6.92 & 67.79 & 53.01 & 48.51 & 50.15 \\
        \midrule
        Gate Pruning & 550.27 & 496.34 & $>$1000 & $>$1000 & 29.45 & 26.17 & 24.51 & 24.75 \\
        Up Pruning & 18.57 & 63.59 & 68.15 & 20.44 & 52.09 & 29.94 & 25.32 & 30.37 \\
        DejaVu & 6.82 & 10.55 & 11.25 & 6.83 & 64.77 & 49.41 & 41.5 & 49.38 \\
        CATS & 196.11 & 122.93 & $>$1000 & $>$1000 & 34.8 & 27.89 & 25.25 & 25.12 \\
        CATS+LoRA & 6.90 & 9.56 & 787.70 & 20.33 & 60.05 & 44.67 & 23.63 & 26.31 \\
        \textbf{DIP} & 6.5 & 8.66 & 9.01 & 6.16 & \underline{72.95} & 60.33 & 53.72 & 55.92 \\
        \textbf{DIP+LoRA} & \underline{5.64} & \underline{7.68} & \underline{8.71} & \underline{6.03} & 72.43 & \underline{61.14} & \underline{54.35} & \underline{56.20} \\
        \bottomrule
    \end{tabular}
    \end{small}
\end{table*}

\section{Hyperparameter tuning}

\subsection{Density of up, down and gate matrices}
\label{app:tune_up_down_sparsity}
We determined the optimal memory allocation for DIP in a three steps procedure illustrated in Figure~\ref{fig:up_gate_alloc}:
\begin{enumerate}
    \item Run a 2D optimization on Memory vs Perplexity and determining the pareto optimal configurations
    \item Modelling the pareto optimal solutions in the parameter space. Here we considered linear models from the target MLP density and the density of the up and gate matrices in logit space.
    \item Use the fitted model to determine the optimal allocation for a target MLP density.
\end{enumerate}

Using the same procedure, we determined there is no significant gain in using different sparsity levels for the Up $\Wu$ and Gate $\Wg$ matrices (Figure~\ref{fig:up_gate_alloc}).
We found that the optimal allocation is consistent across the tested LLMs.

\section{Additional Results}
\label{app:additional_results}
\begin{figure*}
    \centering
    \begin{minipage}{0.49\textwidth}
     \includegraphics[width=\textwidth]{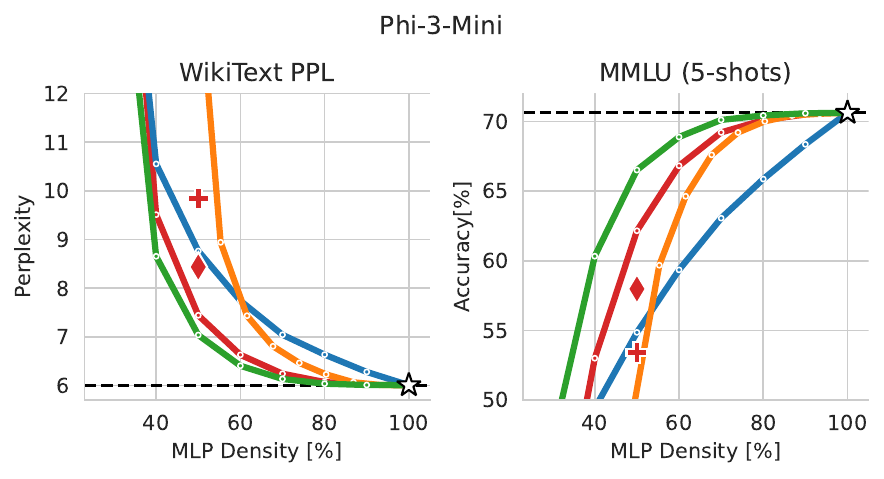}   
    \end{minipage}
    \begin{minipage}{0.49\textwidth}
     \includegraphics[width=\textwidth]{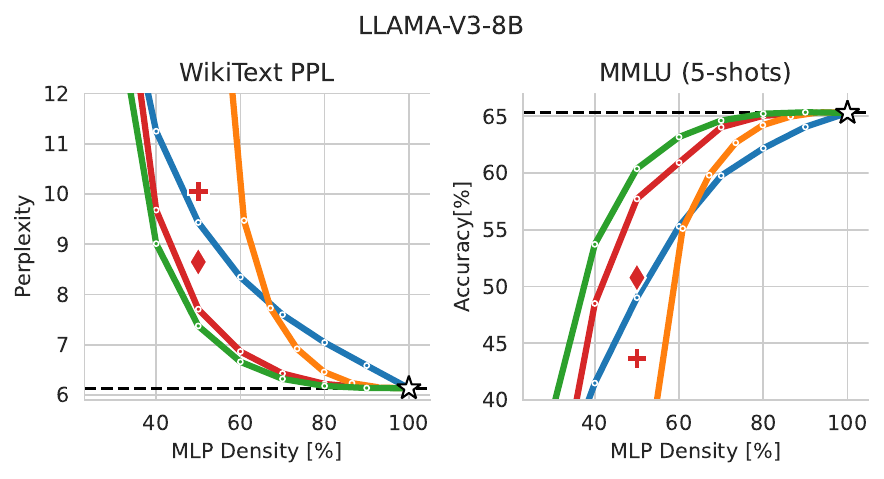}   
    \end{minipage}
    \begin{minipage}{0.65\textwidth}
     \includegraphics[width=\textwidth]{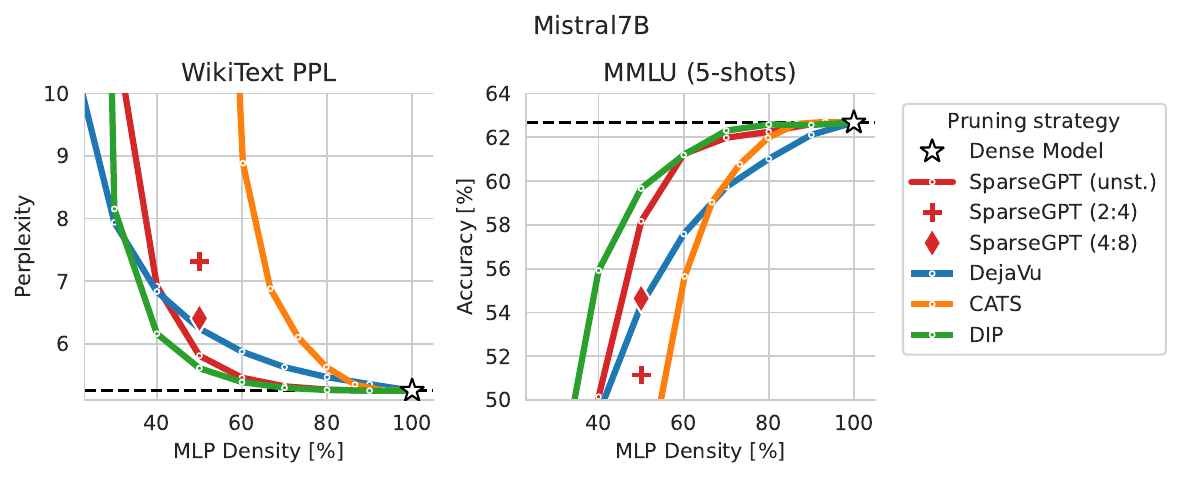}   
    \end{minipage}
    \caption{Downstream WikiText Perplexity and MMLU accuracy as a function of MLP Density for Phi-3-Mini, LLama-v3-8B, and Mistral-7B models.}
    \label{fig:additional_ppl_mmlu_pareto}
\end{figure*}

We show in Figure \ref{fig:additional_ppl_mmlu_pareto} additional results in terms of perplexity and MLP sparsity for Phi-3-Mini, Llama-v3-8B and Mistral-7B, complementing the study on Phi-3-Medium presented in Section 6.2. Table~\ref{tab:exp1_60} and Table~\ref{tab:exp1_40} report the perplexity and accuracy values evaluated at 60\% and 40\% MLP density respectively. Consistently to the results in Table~\ref{tab:exp1}, we don't consider memory overhead introduced by different methods. The MLP density for CATS and CATS+LoRA may vary up to 2\% from the operating point since the fixed estimated threshold may result in slightly different sparsity levels when evaluated on different datasets.

Table~\ref{tab:additional_tasks} compare the performance of the same models at 50\% MLP sparsity on a wide range of tasks including common reasonong (ARC, BoolQ, HellaSwag, PIQA, Wingogrande), multi-lingual reasonong (MGSM) and language understanding (MMLU-Pro). The evaluation procedures follows the protocol described in the \textit{Language Model Evaluation Harness} \citep{eval-harness}.  MGSM and MMLU-Pro require generation of multiple tokens and the corresponding accuracy is evaluated by considering exact matches. The reported values of MGSM refer to the average performance in 11 languages. 

These additional results reinforce the trend we observed for perplexity and MMLU in Table~\ref{tab:exp1}, with DIP outperforming unstructured SparseGPT, DejaVU and CATS on most tasks and architectures.

\begin{table*}
\caption{Accuracy at 50\% MLP sparsity on various language reasoning and understanding tasks.}
\label{tab:additional_tasks}
\footnotesize{
    \setlength{\tabcolsep}{5pt}
    \centering~\\
\begin{tabular}{l|cccccccc}
    \toprule
    
    &\multicolumn{8}{c}{Accuracy $\uparrow$}\\\cmidrule(r){2-9}
     & ARC (easy) &	ARC (challenge)	&BoolQ & HellaSwag & PIQA &Winogrande & MGSM & MMLU-Pro \\\midrule
      \multicolumn{9}{c}{Phi3Med}\\\midrule
     Dense & 85.40 & 61.43 & 88.53 & 65.07 & 80.85 & 76.56 & 32.95 & 52.69\\
GLU pruning (oracle) & 85.31 & 60.58 & 88.75 & 65.09 & 81.39 & 75.37 & 33.24 & 52.19\\\midrule
SparseGPT (unstructured) & \underline{84.05} & 55.38 & \underline{88.93} & 60.83 & \underline{78.78} & \underline{75.22} & 26.80 & 44.13\\\midrule
DejaVu & 83.33 & 54.52 & 87.74 & 59.90 & 77.86 & 74.82 & 20.11 & 37.02 \\
CATS & 80.22 & 54.78 & 86.18 & 60.73 & 74.48 & 66.22 & 13.60 & 38.50\\
DIP & 81.06 & \underline{56.14} & 88.69 & \underline{62.33} & 76.77 & 72.22 & \underline{33.24} & \underline{48.01}\\\midrule
     \multicolumn{9}{c}{Phi3Mini}\\\midrule
     Dense & 81.99 & 53.75 & 85.17 & 59.01 & 80.41 & 73.48 & 32.87 & 44.46\\
GLU pruning (oracle) & 82.24 & 55.12 & 84.68 & 59.15 & 79.71 & 73.64 & 31.85 & 43.42\\\midrule
SparseGPT (unstructured) & 78.28 & 47.44 & 80.61 & 53.92 & 77.26 & \underline{70.64} & 20.22 & 30.30\\\midrule
DejaVu & 75.29 & 45.14 & 74.59 & 43.74 & 76.50 & 62.98 & 6.33 & 24.17 \\
CATS & 75.55 & 45.65 & 65.99 & 52.50 & 72.91 & 62.12 & 6.40 & 21.76 \\
DIP & \underline{79.42} & \underline{53.33} & \underline{83.15} & \underline{57.22} & \underline{77.31} & 68.43 & \underline{26.36} & \underline{36.66}\\\midrule
     \multicolumn{9}{c}{Llama8B}\\\midrule
     Dense & 80.09 & 50.06 & 81.35 & 60.19 & 79.71 & 72.38 & 9.75 & 34.47 \\
     GLU prining (oracle) & 79.38 & 49.83 & 80.52 & 60.21 & 79.82 & 73.24 & 10.07 & 33.69 \\\midrule
     SparseGPT (unstructured) & 75.42 & 41.30 & 78.99 & 54.58 & 77.09 & \underline{71.11} & 5.13 & 24.00\\\midrule
     DejaVu & 67.59 & 39.76 & 74.89 & 53.89 & 76.44 & 68.59 & 3.27 & 20.10\\
     CATS & 54.38 & 32.25 & 56.70 & 45.43 & 70.24 & 57.54	& 0.47 & 7.65\\
     DIP & \underline{77.57} & \underline{46.50} & \underline{79.51} & \underline{57.90} & \underline{77.53} & \underline{71.11} & \underline{7.35} & \underline{29.48}\\\midrule
     \multicolumn{9}{c}{Mistral7B}
     \\\midrule
    Dense & 80.89 & 50.43 & 83.61 & 61.21 & 80.58 & 73.88 & 4.40 & 30.39\\
GLU pruning (oracle) & 80.51 & 50.26 & 83.33 & 61.59 & 80.63 & 73.09 & 4.47 & 30.18\\\midrule
SparseGPT (unstructured) & 76.68 & 44.88 & 76.24 & 57.05 & 78.18 & 71.82 & 4.11 & 24.19\\\midrule
DejaVu & 77.40 & 45.90 & 79.24 & 60.47 & 79.33 & 72.22 & 2.87 & 23.55\\
CATS & 51.56 & 32.00 & 73.85 & 49.67 & 70.02 & 61.64 & 0.47 & 5.66\\
DIP & \underline{79.12} & \underline{48.98} & \underline{83.15} & \underline{60.70} & \underline{79.54} & \underline{72.53} & \underline{4.29} & \underline{28.02}\\\bottomrule
\end{tabular}
}
\end{table*}

\section{Ablation on hardware specifications}
\label{app:ablation_flash}

\begin{table}[h!]
    \centering
    \caption{Comparison of throughput for dynamic sparsity methods at different DRAM sizes with Phi-3-Medium quantized to 4 bits. We report the highest throughput achieved at a 0.5 increase in perplexity on WikiText-2 over the dense model.} 
    \label{tab:hardware_dram}
    \vskip 0.1in
    
    \setlength{\tabcolsep}{12pt}
    \begin{tabularx}{\columnwidth}{l|cccc}  
        \toprule
        DRAM size & 2 GB  & 4 GB & 6 GB\\
        \midrule
        & \multicolumn{3}{c}{Throughput [tok/s] $\uparrow$ } \\
        \midrule
        Dense               & 0.19 & 0.29 & 0.71  \\
        GLU Pruning~~~~~~~  & 0.24 & 0.45 & 1.83  \\
        Up Pruning~~~~~~~  & 0.27 & 0.52 & 1.35  \\
        CATS                & 0.25 & 0.47 & 1.21  \\
        \textbf{DIP-CA}     & \underline{0.31} & \underline{0.56} & \underline{1.94}   \\
        \bottomrule
    \end{tabularx}
\end{table}

For main experiments we considered a fixed hardware setting in line with the specifications of Apple A18 processors. We now consider different target devices and scenarios by simulating changes in DRAM size and Flash read speed.

\paragraph{DRAM size}
In Table \ref{tab:hardware_dram} we consider use-cases with lower DRAM availability at 2GB, simulating the use-case of budget smartphones, and higher DRAM availability at 6GB, which represents high-end devices, or scenarios where the OS and background applications require less memory. In all scenarios, DIP yields better throughput at a fixed perplexity increase of 0.5 over the baseline. Remarkably, DIP improves throughput by 170\% against the dense model with 6GB of DRAM. This stems from the increased availability in caching space for the linear layer, which increases the cache hit rate to 89\%, compared to the 53\% at 4GB and only 8\% at 2GB. At even higher DRAM sizes, where the dense model fully fits in cache, we expect GLU Pruning to outperform DIP-CA, as the overhead in loading more weights from DRAM (for GLU Pruning) has a minor impact with respect to the loss in accuracy resulting from sparsifying all MLP layers (in DIP-CA).

\begin{table}[h!]
    \centering
    \caption{Comparison of throughput for dynamic sparsity methods at different Flash reading speeds with Phi-3-Medium quantized to 4 bits. We report the highest throughput achieved at a 0.5 increase in perplexity on WikiText-2 over the dense model.} 
    \label{tab:hardware_flash}
    \vskip 0.1in
    
    \setlength{\tabcolsep}{10pt}
        \begin{tabularx}{\columnwidth}{l|cccc}  
        \toprule
        Flash read speed & 0.5 GB/s & 1 GB/s & 2 GB/s\\
        \midrule
        & \multicolumn{3}{c}{Throughput [tok/s] $\uparrow$ } \\
        \midrule
        Dense               & 0.15 & 0.29 & 0.59  \\
        GLU Pruning~~~~~~~  & 0.23 & 0.45 & 0.91  \\
        Up Pruning~~~~~~~  & 0.26 & 0.52 & 1.01  \\
        CATS                & 0.24 & 0.47 & 0.91  \\
        \textbf{DIP-CA}     & \underline{0.28} & \underline{0.56} & \underline{1.09}   \\
        \bottomrule
    \end{tabularx}
\end{table}

\paragraph{Flash reading speed}
We evaluate DIP against previous methods at different operating points in terms of Flash reading speed. The actual transfer speed can change significantly depending on storage type, hardware interfaces, and data representation for the saved models. While takeaways and relative improvements at different reading speeds do not change, the results in Table \ref{tab:hardware_flash} highlight how the absolute throughput values change almost at the same rate as the increases in Flash reading speed, confirming that this is the main bottleneck to achieve high-latency in memory-constrained scenarios.

\input{artifact-appendix}

\end{document}

%% file: math.tex
\def\W{\mathbf{W}}

\def\xt{\mathbf{x}^{(t)}}
\def\ct{\mathbf{c}^{(t-1)}}
\def\ctt{\mathbf{c}^{(t)}}
\def\mt{\mathbf{m}^{(t)}}
\def\st{\mathbf{s}^{(t)}}

\def\update{\operatorname{cache\_update}}
\def\topk{\operatorname{topK}}
\def\Wu{\bm{W}^{u}}
\def\Wg{\bm{W}^{g}}
\def\Wd{\bm{W}^{d}}
\def\aW{\widetilde{\bm{W}}}
\def\aWu{\widetilde{\bm{W}}^{u}}
\def\aWg{\widetilde{\bm{W}}^{g}}
\def\aWd{\widetilde{\bm{W}}^{d}}
\def\MLP{\text{MLP}}
\def\GLU{\text{GLU}}
\def\aMLP{\widetilde{\text{MLP}}}
\def\aGLU{\widetilde{\text{GLU}}}
\def\pred{{pred}}

\def\rvx{{\mathbf{x}}}

\def\mA{{\bm{A}}}
\def\mB{{\bm{B}}}

\def\mW{{\bm{W}}}

\DeclareMathAlphabet{\mathsfit}{\encodingdefault}{\sfdefault}{m}{sl}
\SetMathAlphabet{\mathsfit}{bold}{\encodingdefault}{\sfdefault}{bx}{n}



%% file: artifact-appendix.tex
\section{Artifact Appendix}

\subsection{Abstract}

This Artifact Appendix describes the experimental workflow, artifacts and results from this paper submitted for the Artifact Evaluation at MLSys 2025. The experiments described in this paper can be reproduced through the open source code released as a code artifact \cite{federici_2025_15088634}. The code should be executed in a Python 3 environment as described in this appendix. The setup relies on GPU hardware to accelerate the experiments, with the smallest model requiring 40GB of VRAM. Datasets and models used in this paper can be obtained from Huggingface\footnote{\href{https://huggingface.co/models}{https://huggingface.co/models}}. 


\subsection{Artifact check-list (meta-information)}


{\small
\begin{itemize}
  \item {\bf Algorithm: } Efficient LLM inference.
  \item {\bf Models: } Phi-3-Medium, Phi-3-Mini, Llama-v3-8B, Mistral-v01-7B.
  \item {\bf Data sets: } Wikitext, MMLU, SlimPajama (optional).
  \item {\bf Run-time environment: } Python 3.8 and above interpreter and latest pip version to install the required packages.
  \item {\bf Hardware: } A torch-enabled GPU with 80GB of VRAM is required for most models. For Phi-3-Mini, 40GB of VRAM are enough.
  \item {\bf Execution: } \textit{python scripts/run\_experiment.py --help}
  \item {\bf Metrics: } Perplexity, 5-shot accuracy, throughput, memory footprint, MLP density.
  \item {\bf Output: } Metrics for a given method, model, dataset and hardware simulator setup.
  \item {\bf Experiments: } Experiments to reproduce Table \ref{tab:exp1} (perplexity and 5-shot accuracy at 50\% sparsity) and Table \ref{tab:exp2} (throughput at given perplexity increase) in the paper.
  \item {\bf How much disk space required (approximately)?: } Less than 1 MB for the code, Up to 1GB for the pyhon environment setup. Up to 100GB for the LLM models.
  \item {\bf How much time is needed to prepare workflow (approximately)?: } A few minutes for the environment setup and download each model and dataset.
  \item {\bf How much time is needed to complete experiments (approximately)?: } Less than 10 minutes for a single run with Phi-3-Mini on Wikitext. More than 200 GPU hours to reproduce all experiments.
  \item {\bf Publicly available?: } Yes.
  \item {\bf Code licenses (if publicly available)?: } BSD 3-Clause Clear License.
  \item {\bf Data licenses (if publicly available)?: } As specified for each Huggingface dataset.
  \item {\bf Workflow framework used?: } No.
  \item {\bf Archived (provide DOI)?: } \href{https://doi.org/10.5281/zenodo.15088634}{10.5281/zenodo.15088634}
\end{itemize}
}

\subsection{Description}

\subsubsection{How delivered}

Our source code and instructions for installation, setup and experiment reproduction are publicly available at: \href{https://github.com/Qualcomm-AI-research/dynamic-sparsity}{https://github.com/Qualcomm-AI-research/dynamic-sparsity}.

\subsubsection{Hardware dependencies}
We recommend testing on a Linux machine with torch-enabled GPU. We used a x86\_64 architecture with an AMD 64-core processor, Nvidia H100 with 80GB of VRAM and CUDA 12.2.

\subsubsection{Software dependencies}
Python 3.8 and above interpreter and latest pip version to install the required packages. Tested with Python 3.8 and pip 24.3.1. Packages are listed in \textit{\href{https://github.com/Qualcomm-AI-research/dynamic-sparsity/blob/main/requirements.txt}{requirements.txt}}.

\subsubsection{Data sets}
The Wikitext and MMLU data sets are required to reproduce the main experiments, and SlimPajama is needed when running the method variation employing LoRA finetuning.

\subsection{Installation}

Python 3 environment with installed packages as listed in \textit{\href{https://github.com/Qualcomm-AI-research/dynamic-sparsity/blob/main/requirements.txt}{requirements.txt}}. Dataset and models must be downloaded from Huggingface. All details on installation and setup are provided in the \enquote{\href{https://github.com/Qualcomm-AI-research/dynamic-sparsity/blob/main/README.md\#getting-started}{Getting Started}} section of the README documentation.

\subsection{Experiment workflow}
In the \enquote{\href{https://github.com/Qualcomm-AI-research/dynamic-sparsity/blob/main/README.md\#usage}{Usage}} subsection of the README file we explain how to run experiments with the released code, including a description of all the main parameters regulating an experiment.

\subsection{Evaluation and expected result}
In the \enquote{\href{https://github.com/Qualcomm-AI-research/dynamic-sparsity/tree/main?tab=readme-ov-file\#reproducing-results}{Reproducing Results}} section, we explain how to prepare and launch the experiments to reproduce the results in Table \ref{tab:exp1} (perplexity and 5-shot accuracy at 50\% sparsity) and Table \ref{tab:exp2} (throughput at given perplexity increase) in the paper.

\subsection{Experiment customization}
The provided source code implements a set of dynamic sparsity methods for efficient LLM inference. We showcase in our paper experimental results on a selected set of data sets and models, but experimenting with different models and data sets is also possible with minimal code changes. Similarly, we release in our source code a hardware simulator to assess the system throughput under different configurations. These configurations are flexible and can be easily extended to simulate new scenarios. As an example, we included in \textit{\href{https://github.com/Qualcomm-AI-research/dynamic-sparsity/tree/main/scripts/config/hw\_simulator/processor}{/scripts/config/hw\_simulator/processor}} the configurations for a variety of Apple processors based on the specifications provided in \href{https://en.wikipedia.org/wiki/Apple_silicon\#Comparison\_of\_A\_series\_processors}{Wikipedia}. An overview of the main components in our codebase is included in the \enquote{\href{https://github.com/Qualcomm-AI-research/dynamic-sparsity/tree/main?tab=readme-ov-file\#repository-structure}{Repository Structure}} section.

